\documentclass[conference]{IEEEtran}
\IEEEoverridecommandlockouts

\usepackage{bm}
\usepackage{booktabs}
\usepackage{makecell}
\usepackage{bbding}
\usepackage{cite}
\usepackage{amsmath,amssymb,amsfonts}
\usepackage{algorithmic}
\usepackage{graphicx}
\usepackage{textcomp}
\usepackage{xcolor}
\usepackage{multirow}
\def\BibTeX{{\rm B\kern-.05em{\sc i\kern-.025em b}\kern-.08em
    T\kern-.1667em\lower.7ex\hbox{E}\kern-.125emX}}
\begin{document}

\title{Anomaly Detection with Prototype-Guided Discriminative Latent Embeddings
}

\author{\IEEEauthorblockN{Yuandu Lai}
\IEEEauthorblockA{\textit{College of Intelligence and Computing} \\
\textit{Tianjin University}\\
Tianjin, China \\
yuandulai@tju.edu.cn}
\and
\IEEEauthorblockN{Yahong Han}
\IEEEauthorblockA{\textit{College of Intelligence and Computing} \\
\textit{Tianjin University}\\
Tianjin, China \\
yahong@tju.edu.cn}
\and
\IEEEauthorblockN{Yaowei Wang}
\IEEEauthorblockA{\textit{Peng Cheng Laboratory} \\
Shenzhen, China \\
wangyw@pcl.ac.cn}
}

\maketitle

\begin{abstract}
Recent efforts towards video anomaly detection (VAD) try to learn a deep autoencoder to describe normal event patterns with small reconstruction errors. The video inputs with large reconstruction errors are regarded as anomalies at the test time. However, these methods sometimes reconstruct abnormal inputs well because of the powerful generalization ability of deep autoencoder. To address this problem, we present a novel approach for anomaly detection, which utilizes discriminative prototypes of normal data to reconstruct video frames. In this way, the model will favor the reconstruction of normal events and distort the reconstruction of abnormal events. Specifically, we use a prototype-guided memory module to perform discriminative latent embedding. We introduce a new discriminative criterion for the memory module, as well as a loss function correspondingly, which can encourage memory items to record the representative embeddings of normal data, i.e. prototypes. Besides, we design a novel two-branch autoencoder, which is composed of a future frame prediction network and an RGB difference generation network that share the same encoder. The stacked RGB difference contains motion information just like optical flow, so our model can learn temporal regularity. We evaluate the effectiveness of our method on three benchmark datasets and experimental results demonstrate the proposed method outperforms the state-of-the-art.
\end{abstract}


\section{Introduction}
\begin{figure}
\centering
\includegraphics[width=\linewidth]{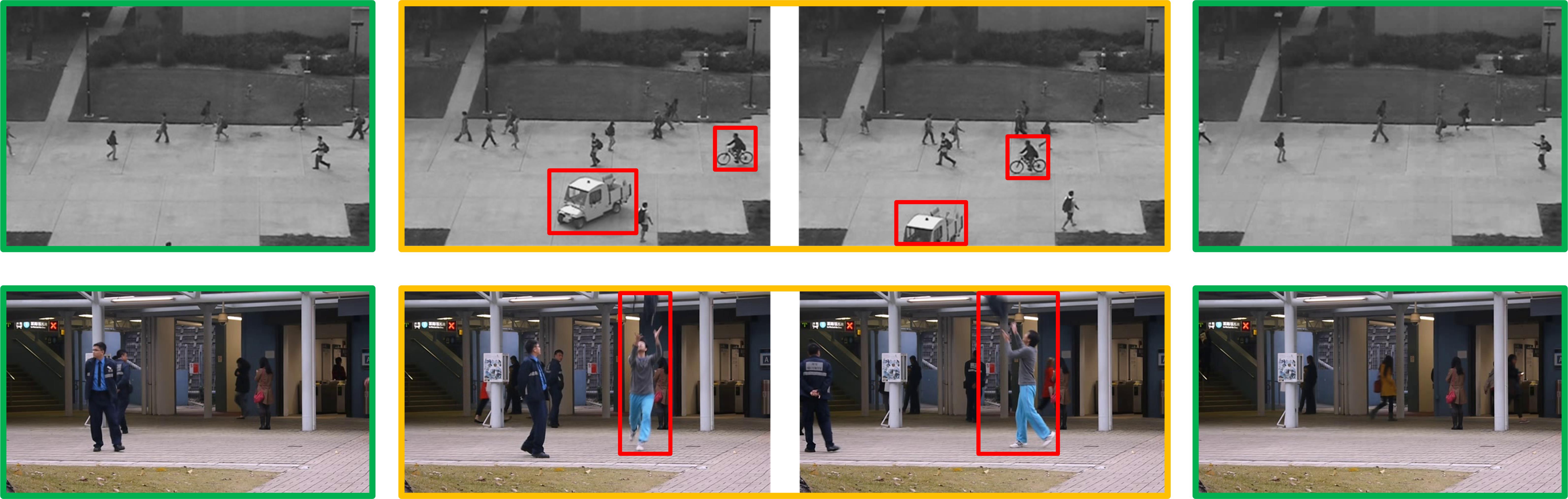}
\caption{Overview of anomaly detection in video sequences. Green and orange rectangles represent normal and abnormal frames, respectively. The red boxes denote the abnormal locations in the pictures. The goal of frame-level anomaly detection is to determine which frames in a video sequence contain anomalies.}
\label{fig1}
\end{figure}

Video anomaly detection (VAD) (as shown in Fig. \ref{fig1}) is widely used in the automatic analysis of surveillance videos, such as traffic, airport, and station monitoring \cite{xu2018dual,li2020multi}. Due to the scarcity and ambiguity of abnormal event samples, VAD is still a challenging task. Most recent efforts towards VAD \cite{liu2018future,nguyen2019anomaly,park2020learning} are based on deep autoencoder (AE) \cite{kingma2013auto} with an unsupervised setting. In the training phase, only normal video frames are input into the model. They usually extract general feature representations and then attempt to reconstruct the inputs again. While during the testing phase, the new sample with a large reconstruction error is more likely to be an anomaly, i.e., a typical process of anomaly detection.

There are two main problems with these methods. In the first place, they usually extract general feature representations rather than the representative ones of normal data. As a result, AEs can sometimes reconstruct abnormal inputs from general feature representation because of their powerful generalization ability \cite{zong2018deep}. In the next place, some methods do not consider the intrinsical temporal characteristics of video events or capture motion information through optical flow, which is a high computational cost.

To address the first problem, some VAD methods based on autoencoder try to learn more typical latent features from normal data. Gong \emph{et al.} \cite{gong2019memorizing} proposed to augment the autoencoder with a memory module and developed an improved autoencoder called MemAE. They used a sparse addressing strategy to force the memory module to record prototypical normal patterns. Park \emph{et al.} \cite{park2020learning} proposed feature compactness and separateness losses to further improve MemAE,  where individual memory items in the memory correspond to prototypical patterns of normal data.
However, their method lacks the ability to model motion information, so it is unable to capture the temporal regularity.

\begin{figure*}
\centering
\includegraphics[width=1.0\linewidth]{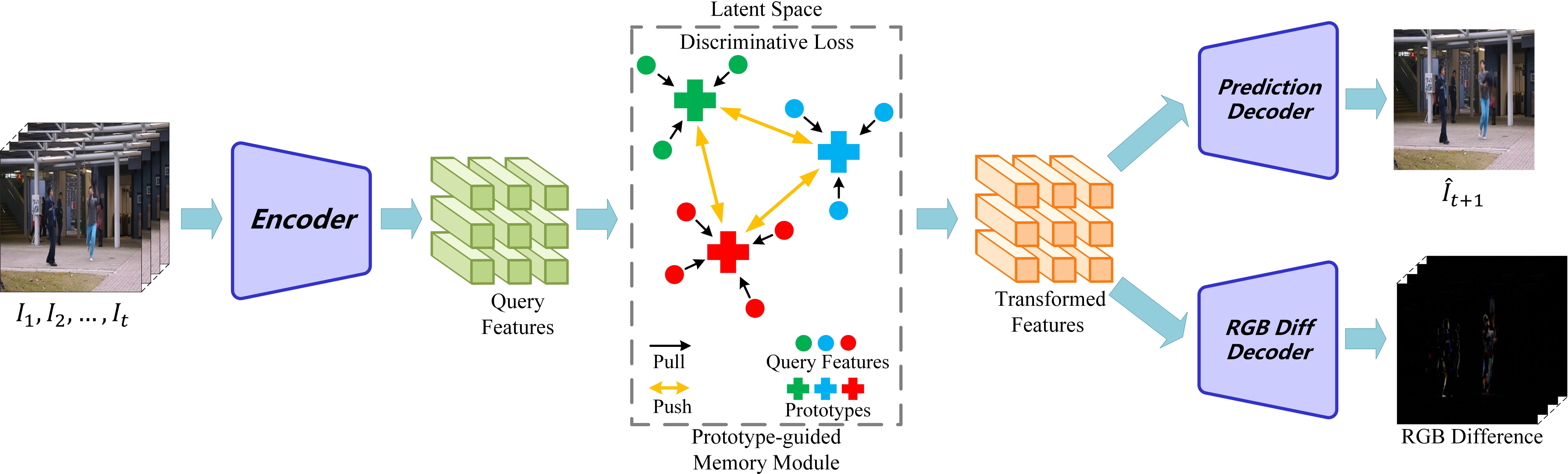}
\caption{Illustration of the proposed video anomaly detection framework. The encoder extracts the query features from an input video clip to retrieve prototypes in the memory. Our discriminative loss encourages prototypes far away from each other and makes the query features close to their nearest prototypes. In this way, the memory module can learn discriminative latent embeddings. The prototype-guided memory module transforms the query features with the most relevant prototypes. Finally, the two decoders input the transformed features and predict the future frame and generate the stacked RGB difference, respectively.}
\label{fig2}

\end{figure*}

In this paper, we propose a prototype-guided memory module and embed it into our two-branch autoencoder. As shown in Fig. \ref{fig2}, we propose to use a novel memory module to record the representative embeddings of normal data, i.e. prototypes, to the items in the memory. Based on cluster assessment, we introduce a discriminative criterion for memory module, and derive a single loss directly from it. Training the memory module with this loss can make the memory items far away from each other and make the features close to their nearest memory item. As a result, these items not only encode representative embeddings of normal data but also preserve the variances. We regard such items as discriminative prototypes. Given a query feature, our memory module will find the most relevant prototypes based on the cosine similarity between the query feature and all memory items. Then we transform the query feature by a weighted average of these prototypes. In this way, even if it is an abnormal input, the feature will tend to be normal after transformation, which prevents the model from reconstructing the abnormal input.

Besides, video events intrinsically own spatio-temporal characteristics, so the appearance and motion features of video events are very crucial for video analysis. Nguyen \emph{et al.} \cite{nguyen2019anomaly} proposed to combine a reconstruction network and an optical flow prediction network that share the same encoder to learn appearance-motion correspondence. However, training their model requires an auxiliary network \cite{dosovitskiy2015flownet} to provide ground truth optical flow, and it is a high computational cost. Chang \emph{et al.} \cite{chang2020clustering} showed that using an RGB difference strategy \cite{wang2018temporal} to simulate motion information is faster than optical flow. Based on this, we propose to add a decoder branch to the traditional video frame appearance reconstruction autoencoder, which is used to learn motion information. As shown in Fig. \ref{fig2}, we design a motion decoder to generate the stacked RGB difference, which helps the model to detect the anomalies related to fast moving in the surveillance videos. This design makes our method more suitable for the task of real-time video anomaly detection. We perform comprehensive experiments on three benchmarks, namely UCSD Ped2 \cite{mahadevan2010anomaly}, CUHK Avenue \cite{lu2013abnormal}, and ShanghaiTech \cite{luo2017revisit} to verify the effectiveness of the proposed method.

Our main contributions of this paper could be summarized as follows.

(1) We devise a novel approach for anomaly detection, which utilizes discriminative prototypes of normal data to reconstruct video frames. A new prototype-guided memory module is designed to perform discriminative latent embedding. We propose a novel discriminative criterion for memory module, as well as a loss function correspondingly, which can encourage memory to record the discriminative prototypes.

(2) We develop a novel two-branch autoencoder, i.e. a future frame prediction branch and an RGB difference generation branch, in which the two branches are responsible for constraining the encoder to extract the spatial and temporal features from the video clips, respectively.

(3) Experimental results on three benchmark datasets show the proposed method achieves state-of-the-art performance. An extensive ablation study shows the effectiveness of the major components of the proposed method.

\section{Related Work}
\subsection{Video Anomaly Detection}
The early works on VAD \cite{adam2008robust,antic2011video,li2013anomaly,saligrama2012video,mehran2009abnormal,wu2010chaotic} relied on hand-crafted appearance and motion features, but these methods lack robustness in complex or crowded scenes with occlusion and shadow. Adam \emph{et al.} \cite{adam2008robust} design an algorithm based on multiple local monitors which collect low-level statistics. The anomaly detector proposed by Li \emph{et al.} \cite{li2013anomaly} is based on a video representation that accounts for both appearance and dynamics, using a set of mixture of dynamic textures models. Saligrama \emph{et al.} \cite{saligrama2012video} develop a probabilistic framework to account for such local spatio-temporal anomalies. A key insight of this work is that if anomalies are local optimal decision rules are local even when the nominal behavior exhibits global spatial and temporal statistical dependencies. \cite{wu2010chaotic} is a unique utilization of particle trajectories for modeling crowded scenes, in which they propose new and efficient representative trajectories for modeling arbitrarily complicated crowd flows.

Recently, deep learning has been successful in many computer vision tasks \cite{krizhevsky2012imagenet,long2015fully,ren2016faster} as well as anomaly detection \cite{sultani2018real}. Due to the scarcity and diversity of anomalies, it is difficult to collect sufficient and comprehensive anomaly data as the supervision information for training model. Thus most existing researches regard VAD as an unsupervised learning task \cite{zhao2017spatio,liu2018future,liu2019margin}, that is to say, only normal data can be obtained during training. Many anomaly detection methods \cite{hasan2016learning,zhao2017spatio,liu2018future} attempted to learn normal event patterns by a deep AE. Xu \emph{et al.} \cite{xu2015learning} proposed the use of stacked denoising autoencoders to separately learn both appearance and motion features as well as a joint representation. Hasan \emph{et al.} \cite{hasan2016learning} use a convolutional autoencoder (ConvAE) to reconstruct video frames, and employed reconstruction error as a criterion to identify anomalies. Based on the LSTM Encoder-Decoder and the Convolutional Autoencoder, Wang \emph{et al.} \cite{wang2018abnormal} explore a hybrid autoencoder architecture, which can better explain the temporal evolution of spatial features. Zhao \emph{et al.} \cite{zhao2017spatio} propose to use a spatio-temporal autoencoder to extract features from both spatial and temporal dimensions. Luo \emph{et al.} \cite{luo2017remembering} leverage a Convolutional LSTMs Auto-Encoder (ConvLSTM-AE) to model normal appearance and motion patterns at the same time,
which further boosts the performance of the ConvAE based solution. There have also been recent attempts to use feedforward convolutional networks for efficient video prediction by minimizing the mean square error (MSE) between the predicted frame and the future frame \cite{mathieu2015deep}. Similar efforts were performed in \cite{liu2018future} using an autoencoder with adversarial training strategy to predict future frame rather than the input itself. Liu \emph{et al.} \cite{liu2018future} believed that normal events are predictable while abnormal ones are unpredictable. However, these methods sometimes even reconstruct or predict abnormal frames well because of the powerful generalization ability of the autoencoder. To alleviate this issue, we apply a new prototype-guided memory module to lessen the representation capacity of autoencoder.

The method most relevant to us is proposed by \cite{nguyen2019anomaly}. Their method learns appearance-motion correspondence by combining a reconstruction network and an optical flow prediction network. In contrast, we design a motion decoder to generate the stacked RGB difference, which helps detect the anomalies related to fast moving. Besides, because the generation of RGB difference is much faster than optical flow, our method is more suitable for real-time VAD.


\subsection{Latent Representation Learning}
Recently, some works have considered the task of unsupervised extraction of meaningful latent representations. Yin \emph{et al.} \cite{yin2020shared} propose a novel multi-view clustering method by learning a shared generative latent representation that obeys a mixture of Gaussian distributions. The motivation is based on the fact that the multi-view data share a common latent embedding despite the diversity among the various views. Ye \emph{et al.} \cite{ye2020probabilistic} propose a probabilistic structural latent representation (PSLR), which incorporates an adaptable softmax embedding to approximate the positive concentrated and negative instance separated properties in the graph latent space.

In order to make the model more sensitive to anomalies, some anomaly detection methods \cite{gong2019memorizing,park2020learning,chang2020clustering} based on autoencoder attempt to learn the representative latent features from normal data.
Recently, memory module have been used in anomaly detection for understanding the typical normality. Gong \emph{et al.} \cite{gong2019memorizing} propose to embed a memory module in the latent space of autoencoder. They use a sparse addressing strategy to force the memory module to record prototypical normal patterns. Park \emph{et al.} \cite{park2020learning} further enhance the effectiveness of memory module in VAD task by using feature compactness and separateness losses to train memory. In this way, their memory items can be discriminative. Chang \emph{et al.} \cite{chang2020clustering} design a k-means cluster to force the autoencoder network to generate compact latent representations. Unlike these methods, our model uses a prototype-guided memory module to perform discriminative latent embedding. In this way, our model can learn various and prototypical features of normal data.

\begin{figure}
\centering
\includegraphics[width=1.0\linewidth]{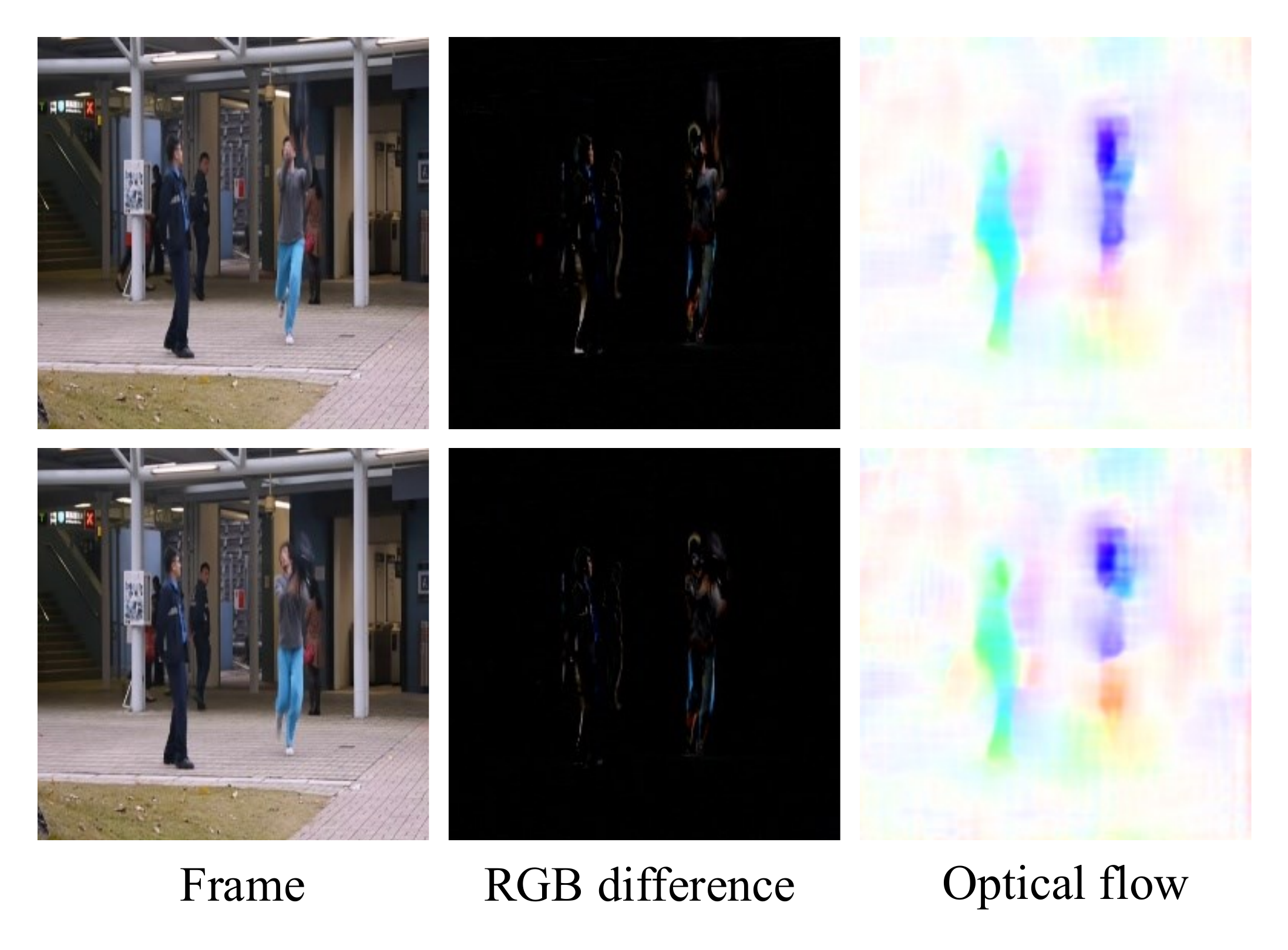}
\caption{Some examples of RGB video frames, RGB difference and optical flow.}
\label{fig-of}
\end{figure}

\section{Proposed Method}
The proposed method leverages a future frame prediction framework for unsupervised anomaly detection. Note that prediction can be considered as a reconstruction of the future frame using previous ones \cite{park2020learning}. Our framework consists of three major components: an encoder (for extracting the query features from video clips), a prototype-guided memory module (for recording discriminative prototypes and transforming the query features), and two decoders (one for decoding the latent features into a future frame, the other for generating RGB difference). The whole framework is illustrated in Fig. \ref{fig2}. Given a video clip with normal event of length $t + 1$. We input the first $t$ frames into the prediction-based network mentioned above, and then the model predicts frame $t + 1$ and generates the corresponding stacked RGB difference. As shown in Fig. \ref{fig-of}, the motion representation learned by RGB difference can simulate the motion information captured by optical flow. We train our model using intensity loss, RGB difference loss, and discriminative loss end-to-end. The memory items are updated to record discriminative prototypes of normal data in the training phase. In the test phase, the model inputs the features transformed by the prototypes to perform future frame prediction. In this way, the model will favor the prediction of normal events and distort the prediction of abnormal events, so as to carry out anomaly detection.
We provide the details of our method as follows.

\subsection{Two-Branch Autoencoder}
The proposed framework is a novel two-branch autoencoder, which can simultaneously learn appearance and motion information from video input. For the shared encoder and future frame prediction decoder, we adopt the U-Net \cite{ronneberger2015u} architecture, which is widely used for the tasks of future frame prediction \cite{liu2018future,park2020learning}. The RGB difference generation decoder of our model keeps the same structure as the future frame prediction decoder except for the skip connections. Such skip connections in U-Net has been proved to be useful for future frame prediction because it can transform low-level features from original domains to the decoded ones. But it is harmful to produce RGB difference, because the network may allow input information to go through these connections directly, resulting in merely copying the input frames.

Mathematically, given a video clip $\bm{x}_{clips}$ with consecutive $t$ frames $I_1, I_2,. . ., I_t$, we sequentially stack all these frames and input them into the encoder to get an encoded representation $\bm{z}$ of size $H \times W \times C$, where $H$, $W$, $C$ denote height, width, and the number of channels, respectively. The encoded representation is used as query features to retrieve the related items in the memory. The memory module transforms the query features $\bm{z}$ with the most relevant prototypes.
\begin{equation}
\bm{z} = f_e(\bm{x}_{clips}; \theta_e),
\end{equation}

\begin{equation}
\widehat{\bm{z}} = f_m(\bm{z}; \theta_m),
\end{equation}
where $\theta_e$ and $\theta_m$ denote the parameters of the encoder $f_e(\cdot)$ and the memory module $f_m(\cdot)$, respectively. And $\widehat{\bm{z}}$ represents the transformed features, which have the same shape as $\bm{z}$. A detailed description of the memory module will be given in the next section. Finally, the two decoders input the transformed features and predict the future frame $\widehat{I}_{t+1}$ and generate RGB difference $\widehat{\bm{x}}_{diff}$, respectively.

\begin{equation}
\widehat{I}_{t+1} = f_d^{pred}(\widehat{\bm{z}}; \theta_d^{pred}),
\end{equation}

\begin{equation}
\widehat{\bm{x}}_{diff} = f_d^{diff}(\widehat{\bm{z}}; \theta_d^{diff}),
\end{equation}
where $\theta_d^{pred}$ and $\theta_d^{diff}$ denote the parameters of the future frame prediction decoder $f_d^{pred}(\cdot)$ and the RGB difference generation decoder $f_d^{diff}(\cdot)$, respectively. And we denote the ground truth RGB difference between $\bm{x}_{clips}$ and $I_{t+1}$ as $\bm{x}_{diff}$:
\begin{equation}
\bm{x}_{diff} = \bm{x}_{clips} - I_{t+1}.
\end{equation}
The loss function $\mathcal{L}_{pred}$ and $\mathcal{L}_{diff}$ for the future frame prediction decoder and the RGB difference generation decoder are given in (\ref{eqpred}) and (\ref{eqdiff}):
\begin{equation}\label{eqpred}
\mathcal{L}_{pred} = {\lVert I_{t+1} - \widehat{I}_{t+1} \lVert}_2,
\end{equation}
\begin{equation}\label{eqdiff}
\mathcal{L}_{diff} = {\lVert\bm{x}_{diff} - \widehat{\bm{x}}_{diff}\lVert}_2.
\end{equation}

\subsection{Prototype-Guided Memory Module}
In this section, we introduce our key innovations in detail. In order to enable our model to have the ability to record representative embeddings of normal data for understanding the normality, we propose a prototype-guided memory module and embed it into our two-branch autoencoder. Such a memory module can effectively prevent the model from reconstructing abnormal inputs and improve the performance of anomaly detection.

\begin{figure}
\centering
\includegraphics[width=1.0\linewidth]{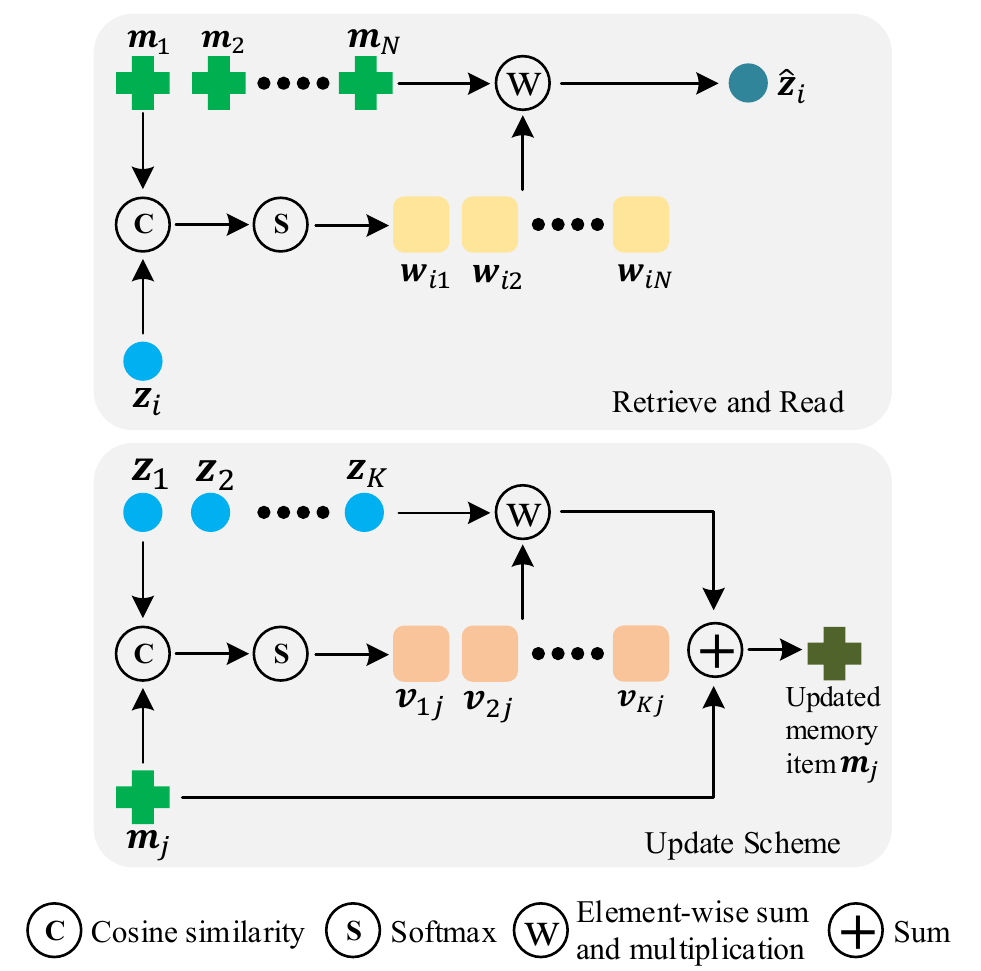}
\caption{The read and update scheme for the memory module.}
\label{fig3-3}
\end{figure}

\subsubsection{Discriminative Criterion}
We consider how to let the memory module learn the representative embeddings of normal data, i.e. prototypes. We argue that the memory module should satisfy the following two points. Firstly, memory module should record various features. This requires that items in the memory module should be as far away from each other as possible in the latent space. If they are close to each other, memory module tend to record similar features, thus lacking discriminability. Secondly, memory module should record prototypical features. This requires the queries to be close to their nearest memory item, reducing intra-class variations. In this way, the memory items are representative and can be regarded as the typical representations of their surrounding features. We refer to these two points as the discriminative criterion of the memory module. In the following, we describe in detail how to read and update memory. These operations are illustrated in Fig. \ref{fig3-3}. The process of deriving the loss function from discriminative criterion will also be given. This discriminative loss will be used to train our model as a term of the objective function.

\subsubsection{Retrieve and Read}
We denote the memory items as a matrix $\bm{M} \in \mathbb{R}^{N \times C}$, where $N$ is the number of memory items and $C$ is the dimension of each memory item. Let row vector $\{\bm{m}_j\}^N_{j=1}$ denotes the $j$-th memory item of $\bm{M}$. And we denote by $\{\bm{z}_i\}^{K}_{i=1}$, where $K = H \times W$, individual queries in the query features $\bm{z}$. Each $\bm{z}_i$ is a query of size $1 \times 1 \times C$, the same shape as $\bm{m}_j$. Queries are input to the memory module to read items. In order to retrieve and read the items, we use (\ref{eq4}) to calculate the cosine similarity between each query $\{\bm{z}_i\}^{K}_{i=1}$ and all memory items $\{\bm{m}_j\}^N_{j=1}$.
\begin{equation}\label{eq4}
d(\bm{z}_i, \bm{m}_j) = \frac{\bm{z}_i\bm{m}_j^\mathsf{T}}{\lVert\bm{z}_i\lVert\lVert\bm{m}_j\lVert}.
\end{equation}
And then we can obtain a two-dimensional distance matrix $\bm{d}$ with a size of $K \times N$. Further, we perform normalization on each row of the matrix $\bm{d}$ by a softmax function as:
\begin{equation}\label{eq5}
\bm{w}_{ij} = \frac{e^{d(\bm{z}_i, \bm{m}_j)}}{\sum_{j=1}^N e^{d(\bm{z}_i, \bm{m}_j)}}.
\end{equation}
With this distance matrix, we can know which memory items are related to query $\bm{z}_i$. We can use (\ref{eq6}) to retrieve and read the memory module to get the transformed feature $\widehat{\bm{z}}_i$.
\begin{equation}\label{eq6}
\widehat{\bm{z}}_{i} = \sum_{j=1}^N \bm{w}_{ij}\bm{m}_j.
\end{equation}
We use all memory items to transform features so that our model can understand various prototypical normal patterns.

\subsubsection{Update Scheme}

To update the memory items, we compute the normalized distance metric between each memory items $\{\bm{m}_j\}^N_{j=1}$ and all queries $\{\bm{z}_i\}^{K}_{i=1}$ as:
\begin{equation}
\bm{v}_{ij} = \frac{e^{d(\bm{z}_i, \bm{m}_j)}}{\sum_{i=1}^K e^{d(\bm{z}_i, \bm{m}_j)}},
\end{equation}
where, $\bm{v}$ is an $K \times N$ distance matrix. We utilize this cosine similarity between memory items and query features to update each memory items using following equation:
\begin{equation}
\bm{m}_j \gets f(\bm{m}_j + \sum_{i=1}^K \bm{v}_{ij}\bm{z}_i),
\end{equation}
where $f(\cdot)$ is the L2 norm. By using the weighted average of query features, we can focus more attention on the query features near the memory item.

\subsubsection{Derivation of Loss Function}
\begin{figure}
\centering
\includegraphics[width=1.0\linewidth]{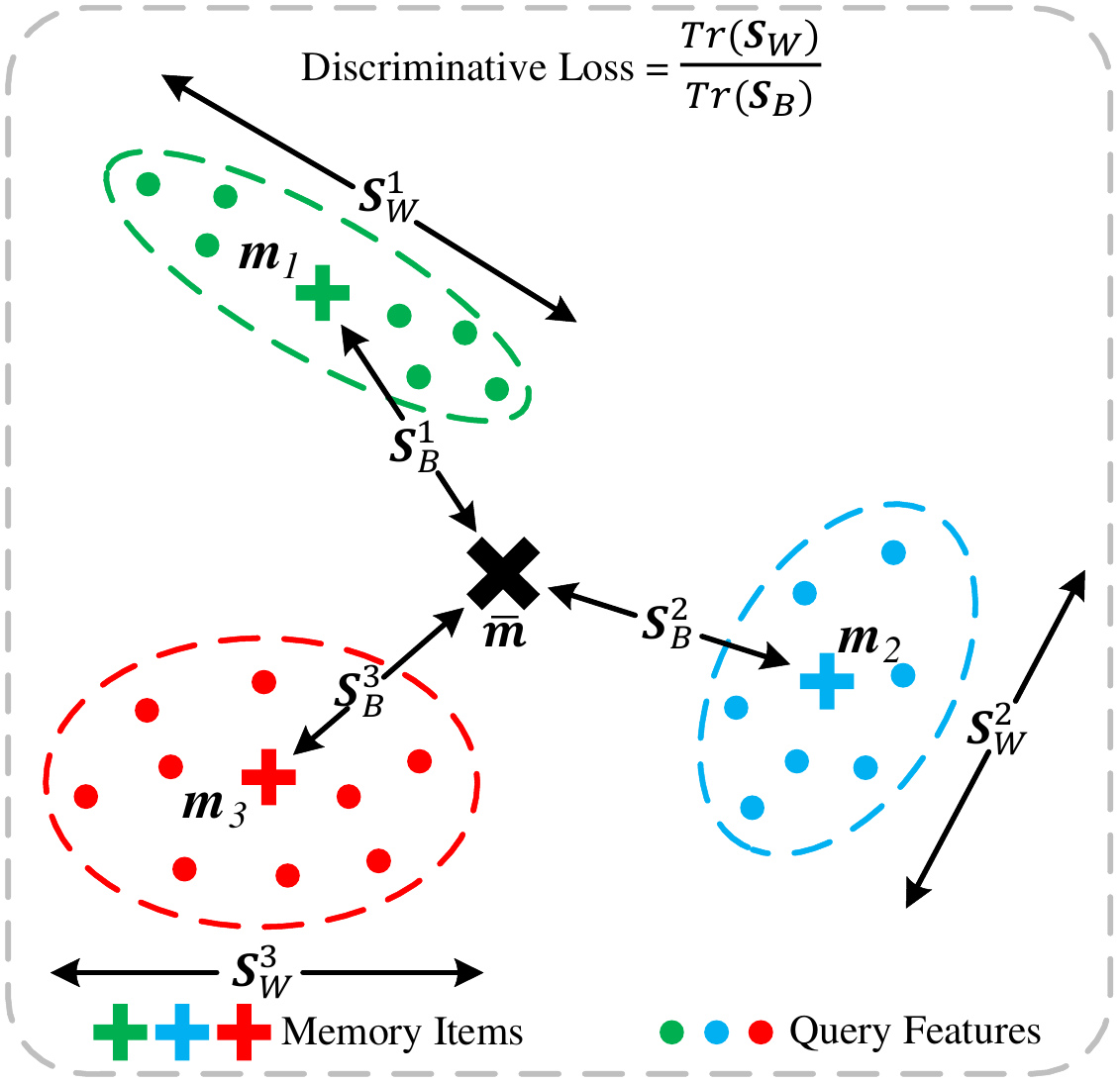}
\caption{The schematic illustration of the proposed discriminative loss. The query features with the same color are classified into the same item. We want the memory items as far away from each other as possible in order to record various features. At the same time, we hope the queries are close to their nearest memory item, reducing intra-class variations. In this way, memory items are encouraged to record representative embeddings.}
\label{fig3}
\vspace{-0.1cm}
\end{figure}

As shown in Fig. \ref{fig3}, we classify queries into their nearest memory item according to the distance matrix $\bm{d}$.  In order to achieve the first point of the discriminative criterion, i.e., memory module should record various features, we want to increase the distance between memory items. We measure the scatter between memory items in terms of the between-class scatter matrix as:
\begin{equation}\label{eq7}
\bm{S}_B = \sum_{j=1}^{N}\frac{n_j}{K}(\bm{m}_j - \bm{\bar{m}})^\mathsf{T}(\bm{m}_j - \bm{\bar{m}}),
\end{equation}
where $n_j$ is the number of queries declared that the item $\bm{m}_j$ is the nearest one and $\bm{\bar{m}}$ is the mean vector of all memory items,
\begin{equation}\label{eq8}
\bm{\bar{m}} = \frac{1}{N}\sum_{j=1}^{N}\bm{m}_j.
\end{equation}
The size of matrix $\bm{S}_B$ is $C \times C$, and its trace $Tr(\bm{S}_B)$ indicates the scatter between memory items. We want to maximize $Tr(\bm{S}_B)$, to encourage memory module to record various features.

As for the second point of the discriminative criterion, i.e., memory module should record prototypical features. This requires the queries to be close to their nearest memory item, reducing intra-class variations. We compute the scatter between them in terms of the within-class scatter matrix as:
\begin{equation}\label{eq9}
\bm{S}_W^j = \sum_{k=1}^{n_j}(\bm{z}_j^k - \bm{m}_j)^\mathsf{T}(\bm{z}_j^k - \bm{m}_j),
\end{equation}
\begin{equation}\label{eq10}
\bm{S}_W = \sum_{j=1}^{N}\frac{n_j}{K}\bm{S}_W^j,
\end{equation}
where $\{\bm{z}_j^k\}_{k = 1}^{n_j}$ denote those queries declared that the item $\bm{m}_j$ is the nearest one. The size of matrix $\bm{S}_W$ is also $C \times C$. We want to minimize $Tr(\bm{S}_W)$, to encourage memory module to record prototypical features.

Based on the above discussion, we use the trace ratio of within-class to between-class scatter as the discriminative loss of our memory module.
\begin{equation}\label{eq11}
\mathcal{L}_{dis} = \frac{Tr(\bm{S}_W)}{Tr(\bm{S}_B)}.
\end{equation}

Our loss encourages the queries to be close to their nearest memory item while separating all memory items to improve discriminability, which improves the ability to record diverse and prototypical normal patterns.

\subsection{Objective Function}

We combine all these losses corresponding to future frame prediction, RGB difference generation, and discriminative memory module, into our objective function:
\begin{equation}\label{eq13}
\mathcal{L} = \lambda_{pred}\mathcal{L}_{pred} + \lambda_{diff}\mathcal{L}_{diff} + \lambda_{dis}\mathcal{L}_{dis},
\end{equation}
where $\lambda_{pred}$, $\lambda_{diff}$, and $\lambda_{dis}$ are the hyper-parameters to balance each losses.

\begin{table}[tbp]
\begin{center}
\caption{AUC of different methods on the UCSD Ped2, CUHK Avenue, and ShanghaiTech datasets. Numbers in bold indicate the best performance and underlined ones are the second best.} \label{tab:cap1}
\begin{tabular}{@{}l|ccc@{}}
\toprule
Methods                                     & UCSD Ped2          & Avenue                   & SH.Tech                       \\ \midrule
MPPCA \cite{kim2009observe}                 & 69.3\%             & N/A                      & N/A                            \\
MPPC+SFA \cite{mahadevan2010anomaly}        & 61.3\%             & N/A                      & N/A                          \\
MDT \cite{mahadevan2010anomaly}             & 82.9\%             & N/A                      & N/A                          \\
DFAD \cite{del2016discriminative}           & N/A                & 78.3\%                   & N/A                         \\ \midrule
ConvAE \cite{hasan2016learning}             & 85.0\%             & 80.0\%                   & 60.9\%                        \\
ConvLSTM-AE \cite{luo2017remembering}       & 88.1\%             & 77.0\%                   & N/A                            \\
AE-Conv3D \cite{zhao2017spatio}             & 91.2\%             & 77.1\%                   & N/A                            \\
Unmasking \cite{tudor2017unmasking}         & 82.2\%             & 80.6\%                   & N/A                             \\
TSC \cite{luo2017revisit}                   & 91.0\%             & 80.6\%                   & 67.9\%                         \\
Stacked RNN \cite{luo2017revisit}           & 92.2\%             & 81.7\%                   & 68.0\%                          \\
Frame-Pred \cite{liu2018future}             & 95.4\%             & 84.9\%                   & 72.8\%                      \\
Siamese Net \cite{ramachandra2020learning}  & 94.0\%             & 87.2\%                   & N/A                           \\
MemAE \cite{gong2019memorizing}             & 94.1\%             & 83.3\%                   & 71.2\%                              \\
AMC \cite{nguyen2019anomaly}                & 96.2\%             & 86.9\%                   & N/A                                 \\
MNAD \cite{park2020learning}                & \underline{97.0\%} & \bf{88.5\%}              & 70.5\%                          \\
CDAE \cite{chang2020clustering}             & 96.5\%             & 86.0\%                   & \underline{73.3\%}               \\ \midrule
Ours                                        & \bf{97.6\%}        & \underline{87.8\%}       & \bf{74.5\%}                      \\ \bottomrule
\end{tabular}
\end{center}
\end{table}

\subsection{Regularity Score}
At test time, following the previous work \cite{park2020learning}, we also calculate the Peak Signal to Noise Ratio (PSNR) between the input frame and its prediction and the distance between the query features and memory items. We use the sum of these two metrics as the regularity score of each frame.

Given a video clip with consecutive $t+1$ frames, we input the first $t$ frames into our prediction-based network, and it will predict a future frame $\hat{I}_{t+1}$. The generation quality of the prediction frame determines whether it is abnormal or not. Following \cite{liu2018future}, we also use PSNR as (\ref{eq14}) for image quality assessment. A higher PSNR of the $t+1$-th frame indicates that it is predicted well. That means it is more likely to be normal.
\begin{equation}\label{eq14}
PSNR(I_{t+1}, \hat{I}_{t+1}) = 10log_{10}\frac{1}{MSE(I_{t+1}, \hat{I}_{t+1})},
\end{equation}
\begin{equation}\label{eq15}
MSE({I}_{t+1}, \hat{I}_{t+1}) = \frac{1}{S^2}\sum_{i=0}^{S}\sum_{j=0}^{S} \lVert {I}_{t+1}(i, j) - \hat{I}_{t+1}(i, j) \lVert_2^2,
\end{equation}
where $i$, $j$ denote the spatial index of a video frame with size of $S \times S$. And then, we use (\ref{eq16}) to normalize PSNR of all frames under a single scene to the range [0, 1].
\begin{equation}\label{eq16}
P(t) = \frac{PSNR(I_{t}, \hat{I}_{t}) - min_tPSNR(I_{t}, \hat{I}_{t})}{max_{t}PSNR(I_{t}, \hat{I}_{t})- min_tPSNR(I_{t}, \hat{I}_{t})}.
\end{equation}
When the frame $I_t$ is abnormal, we obtain a low value of $P(t)$ and vice versa.

\begin{figure}[tbp]
\centering
\includegraphics[width=1.0\linewidth]{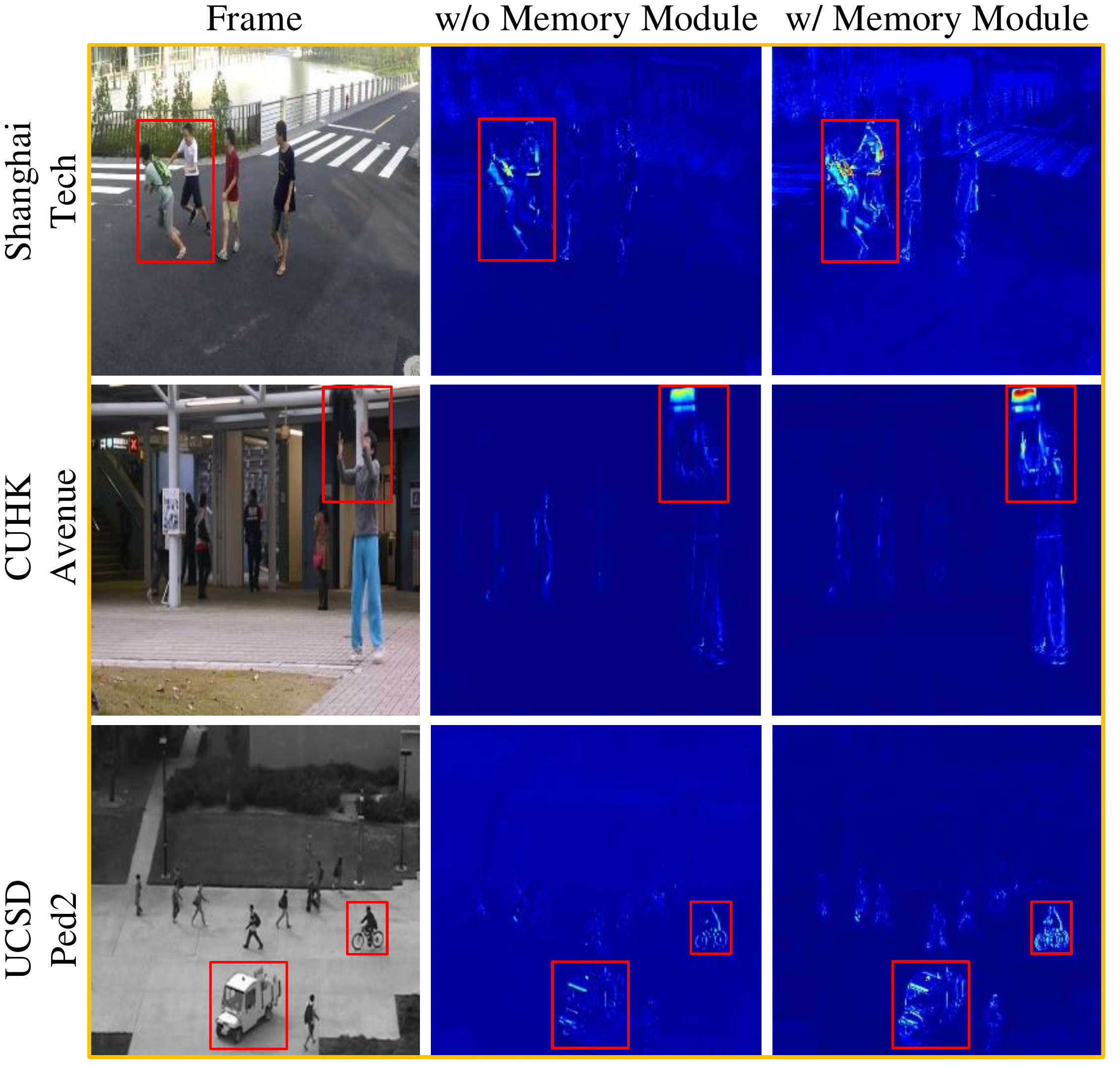}
\caption{Anomaly visualization on ShanghaiTech, CUHK Avenue, and UCSD Ped2 datasets. We show some abnormal frames and their prediction errors. Anomalies are marked with red boxes. Embedding a memory module in the network can detect anomalies more sensitively. Red indicates high anomaly and blue indicates low.}
\label{fig4}
\end{figure}

In addition, the query features from the normal video frame should be close to the memory items in the memory module. We calculate the L2 distance between each query feature and the nearest memory item as another indicator to measure the regularity.
\begin{equation}\label{eq17}
DIST(\bm{z}^t, \bm{m}) = \frac{1}{K}\sum_{i=0}^{K} \lVert \bm{z}_i^t - \bm{m}_{1st} \lVert_2,
\end{equation}
where $1st$ is an index of the nearest memory item for the query feature $\bm{z}_i^t$, defined as:
\begin{equation}
1st = \mathop{\arg\max}_{j \in N} \bm{w}_{ij}.
\end{equation}
And then, we use (\ref{eq18}) to normalize $DIST$ of all frames under a single scene to the range [0, 1].
\begin{equation}\label{eq18}
D(t) = \frac{DIST(\bm{z}^t, \bm{m}) - min_tDIST(\bm{z}^t, \bm{m})}{max_{t}DIST(\bm{z}^t, \bm{m})- min_tDIST(\bm{z}^t, \bm{m})}.
\end{equation}

The value $S(t)$ calculated by (\ref{eq19}) is taken as the regularity score of frame $I_{t}$.
\begin{equation}\label{eq19}
S(t) = \gamma P(t) + (1 - \gamma)(1 - D(t)).
\end{equation}
When the frame $I_t$ is abnormal, we obtain a low value of $S(t)$ and vice versa.

\subsection{Implementation Details}
To train our network, the input images need to be resized to $256 \times 256$ and the intensity of pixels need to be normalized to [-1, 1]. The mini-batch size is 4, and $t$ is set to 4. The size $H \times W \times C$ of query feature map is $32 \times 32 \times 512$. The number of memory items $N$ is normally set to 100. Specifically, our network is trained by Adam optimizer \cite{kingma2014adam} with a learning rate of $2 \times 10^{-4}$. $\lambda_{pred}$, $\lambda_{dis}$, $\lambda_{diff}$, and $\gamma$ normally are 1, 1, 0.1, and 0.6, respectively. Hyper-parameter tuning was done using grid search.

\begin{table}[tbp]
\begin{center}
\caption{Ablation study of our model.} \label{tab:cap2}
\begin{tabular}{@{}c|p{14 pt}<{\centering}p{14 pt}<{\centering}p{14 pt}<{\centering}p{14 pt}<{\centering}p{14 pt}<{\centering}p{14 pt}<{\centering}}
\toprule
\multicolumn{1}{c|}{Component} & \multicolumn{6}{c}{AUC (\%)}     \\ \midrule
RGB difference                    & \XSolidBrush   & \Checkmark     & \XSolidBrush   & \XSolidBrush   & \Checkmark     & \Checkmark   \\
Memory module                     & \XSolidBrush   & \XSolidBrush   & \Checkmark     & \Checkmark     & \Checkmark     & \Checkmark   \\
Discriminative loss               & \XSolidBrush   & \XSolidBrush   & \XSolidBrush   & \Checkmark     & \XSolidBrush   & \Checkmark   \\ \midrule
UCSD Ped2                         & 94.5           & 95.8           & 95.4           & 96.6           & 96.5           & \bf{97.6} \\
CUHK Avenue                       & 84.6           & 85.4           & 85.5           & 86.6           & 86.8           & \bf{87.8}  \\ \bottomrule
\end{tabular}
\end{center}
\end{table}

\subsection{Evaluation Metric}
In order to quantitatively evaluate the effectiveness of our method, following \cite{hasan2016learning,gong2019memorizing,park2020learning}, we also employ frame-level Area Under the Curve (AUC) as the evaluation metric. A threshold on regularity score is varied in order to generate Receiver Operating Characteristic (ROC) curves of false positive rate versus true positive rate. Then the Area Under Curve (AUC) is cumulated to a scalar for performance evaluation. A higher value indicates better anomaly detection performance.

\section{Experiments}
To demonstrate the effectiveness of our method, we conduct experiments on three publicly available datasets and compare our method with different state-of-the-art methods. We also provide an extensive ablation study and some detailed analysis of our proposed method.

\subsection{Datasets}
In this paper, we use three benchmark datasets, including UCSD Ped2 \cite{mahadevan2010anomaly}, CUHK Avenue \cite{lu2013abnormal}, and ShanghaiTech \cite{luo2017revisit}. Now we give a brief introduction to the datasets used in our experiments.

The UCSD Pedestrian dataset contains two subsets: Ped1 and Ped2. Following \cite{gong2019memorizing, nguyen2019anomaly}, we only conduct experiments on Ped2 because Ped1 is frequently used for pixel-wise anomaly detection. The UCSD Ped2 dataset consists of 16 training and 12 testing videos. This dataset contains 12 abnormal events, all of which are unusual pedestrian patterns, such as skateboards, cars, and bicycles, \emph{etc}.

The CUHK Avenue dataset consists of 16 training and 21 testing videos which are captured in CUHK campus avenue. This dataset contains 47 abnormal events, including running, throwing objects, and loitering, \emph{etc}. The size of pedestrian may change with the position and angle of the camera.

 \begin{figure}[tbp]
\centering
\includegraphics[width=0.9\linewidth]{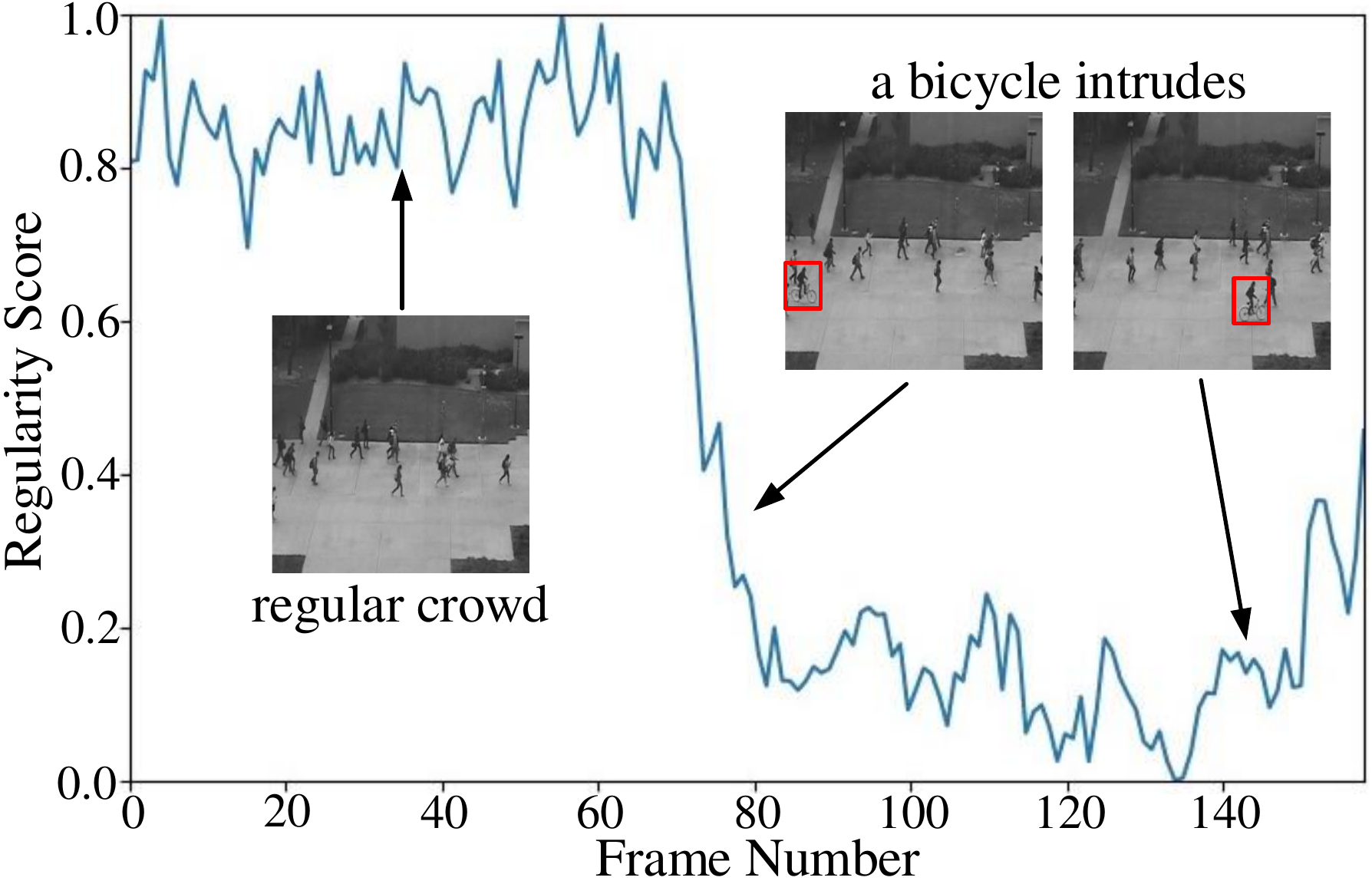}
\caption{Regularity score curve. Some video frames are displayed on the figure, and anomalies are marked with red boxes.}
\label{fig5}
\end{figure}

The ShanghaiTech dataset is the most recent benchmark dataset, which contains 330 training and 107 test videos with 130 abnormal events. It is collected from 13 scenes and is the largest and most challenging anomaly detection dataset at present. Abnormal events are diverse and realistic, which include appearance anomalies like bicycles, motorcycles, cars, skateboards, and motion anomalies such as jumping, chasing, fighting, \emph{etc}.

\subsection{State of the Art Comparison}
Table \ref{tab:cap1} compares our model with some latest deep learning based methods, including ConvAE \cite{hasan2016learning}, ConvLSTM-AE \cite{luo2017remembering}, AE-Conv3D \cite{zhao2017spatio}, Unmasking \cite{tudor2017unmasking}, TSC \cite{luo2017revisit}, Stacked RNN \cite{luo2017revisit}, Frame-Pred \cite{liu2018future}, Siamese Net \cite{ramachandra2020learning}, MemAE \cite{gong2019memorizing}, AMC \cite{nguyen2019anomaly}, MNAD \cite{park2020learning}, and CDAE \cite{chang2020clustering}. In addition, we also compare different hand-craft features based methods, including MPPCA \cite{kim2009observe}, PPC+SFA \cite{mahadevan2010anomaly}, MDT \cite{mahadevan2010anomaly}, and DFAD \cite{del2016discriminative}. Our model achieves the state-of-the-art performance.

Compared to MemAE \cite{gong2019memorizing} which is the first to augment the AE with a memory module to record prototypical patterns of the normal samples, our model uses the loss function derived from discriminative criterion to further enhance the effectiveness of memory module. Our model outperforms MemAE by 3.5\%, 4.5\%, and 3.3\% on UCSD Ped2, ShanghaiTech, and CUHK Avenue, respectively. Compared to AMC \cite{nguyen2019anomaly} which learns appearance-motion correspondence, the performance of our method is better on UCSD Ped2 and CUHK Avenue by large margins. Note that AMC leverages an adversarial learning framework and needs an auxiliary network \cite{dosovitskiy2015flownet} to provide ground truth optical flow. This means that it takes less effort to train our model, but we can still get better performance. Compared to a state-of-the-art prediction method \cite{park2020learning} that also exploits a memory module for anomaly detection, our model outperforms it by 4.0\% on ShanghaiTech and 0.6\% on UCSD Ped2 and has comparable performance on CUHK Avenue. Compared with the latest method \cite{chang2020clustering}, the performance of our model is at least 1.1\% higher than it on all datasets. This demonstrates the effectiveness of our approach to exploiting a two-branch autoencoder with a prototype-guided memory module for video anomaly detection.

\begin{table}[]
\begin{center}
\caption{AUC and running time of different methods on the ped2 dataset. The R and O indicate RGB difference and optical flow, respectively.} \label{tab:cap4}
\begin{tabular}{@{}l|c|c|c@{}}
\toprule
Methods                                    & R/O & AUC    & Running Time \\ \midrule
Frame-Pred \cite{liu2018future}            & O   & 95.4\% & 25 fps       \\
CDAE \cite{chang2020clustering}            & R   & 96.5\% & 32 fps       \\ \midrule
\multirow{2}{*}{Ours}                      & O   & 96.6\% & 30 fps       \\
                                           & R   & 97.6\% & 42 fps       \\ \bottomrule
\end{tabular}
\end{center}
\vspace{-0.3cm}
\end{table}

\subsection{Ablation Study}
To evaluate the effectiveness of each component of our model as shown in Fig. \ref{fig2}, we conduct an extensive ablation study. In Table \ref{tab:cap2}, we measure the performances of variants of our model on UCSD Ped2 and CUHK Avenue. We train the baseline model in the first column with only future frame prediction loss. From the second column, we can see that our model with the RGB difference generation decoder gives better results. This shows that using an RGB difference strategy to simulate motion information can improve performance. The third column indicates the performances of our model with memory module but without discriminative loss. We see better results than the baseline model because the memory module can help the model record normality. The fourth column demonstrates that discriminative loss can further boost memory module performance. Combining all the components can arrive at our complete model in the last column. Benefit from the discriminative memory module and motion information learning, this instance achieves the highest performance.

\subsection{Comparison with Optical Flow}
In order to compare the running time and performance of RGB difference and optical flow, we conducted a comparative experiment on the UCSD Ped2 dataset. Table \ref{tab:cap4} compares our model with some state-of-the-art methods, including Frame-Pred \cite{liu2018future} and CDAE \cite{chang2020clustering}. The R and O indicate RGB difference and optical flow, respectively. We can replace the part of RGB difference in our model with the optical flow method FlowNet \cite{dosovitskiy2015flownet}, which is consistent with the Frame-Pred \cite{liu2018future}. As shown in Table \ref{tab:cap4}, Our method is much faster and performs better than the method based on optical flow. Also based on the RGB difference, our method is faster than CDAE \cite{chang2020clustering}, because the two branch networks of our model share the same encoder, while CDAE \cite{chang2020clustering} contains two different encoders, which increases the running time.

\subsection{Visualization}
In order to further illustrate the effectiveness of our method, we visualize several anomaly detection examples.

\subsubsection{Anomaly Visualization}
As shown in Fig. \ref{fig4}, we select three abnormal samples from datasets for anomaly visualization.
Their corresponding abnormal events are chasing, throwing bag, and vehicle moving on the sidewalk, respectively. The error maps mainly highlight the abnormal locations in the pictures. The error maps in the middle column show that the model can predict anomalies well without our prototype-guided memory module. On the contrary, when the memory module is embedded, the detected anomalies are more significant. This proves that using discriminative prototypes to predict video frames can effectively prevent the model from predicting anomalies.

\subsubsection{Anomaly Event Detection}
In order to show how to detect abnormal events in real-world surveillance videos, we present a regularity score curve for a testing video of UCSD Ped2. As shown in Fig. \ref{fig5}, the anomaly event is a bicycle breaking into the sidewalk. As can be seen from the curve, when there are only pedestrians in the scene, the regularity score maintains a very high value, while when the anomalies occur, the regularity score decreases sharply. This shows that our method is capable of detecting the occurrence of anomalies.

\begin{table}[]
\begin{center}
\caption{AUCs of our model with different memory sizes.} \label{tab:cap3}
\begin{tabular}{@{}c|cc@{}}
\toprule
Memory Size & UCSD Ped2 & CUHK Avenue \\ \midrule
w/o         & 95.8\%    & 85.4\%      \\
25          & 96.6\%    & 86.3\%      \\
50          & 97.0\%    & 86.9\%      \\
100         & 97.6\%    & 87.8\%      \\
200         & 97.5\%    & 87.6\%      \\ \bottomrule
\end{tabular}
\end{center}
\vspace{-0.3cm}
\end{table}

\subsubsection{Predicted Frames Comparison}
\begin{figure*}
\centering
\includegraphics[width=\linewidth]{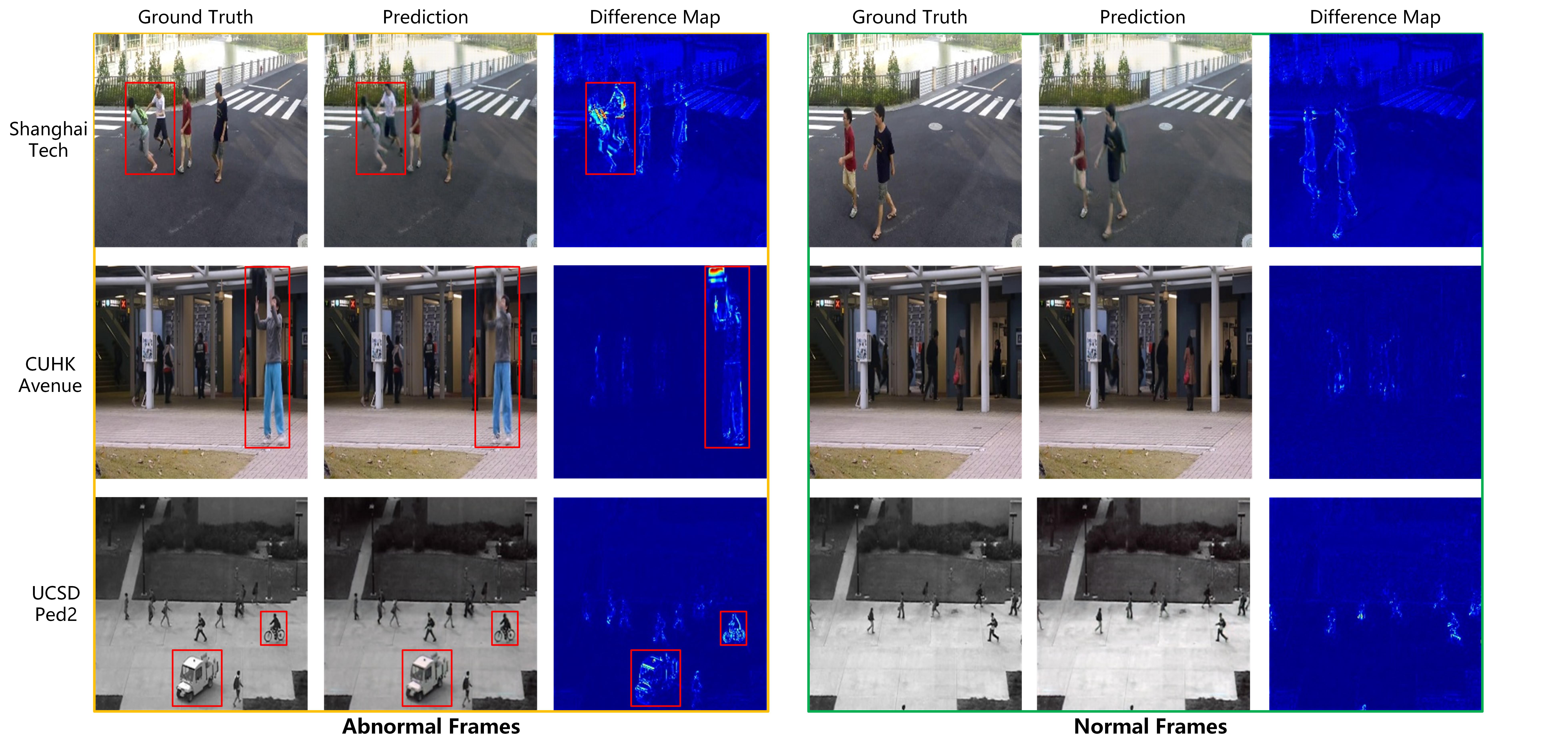}
\caption{Anomaly visualization on ShanghaiTech, CUHK Avenue, and UCSD Ped2 datasets. We show some abnormal and normal frames and their predictions. Anomalies are marked with red boxes. The difference maps mainly highlight the abnormal locations in the picture. Red indicates high anomaly and blue indicates low. Best viewed in color.}
\label{fig4-2}
\vspace{-0.3cm}
\end{figure*}

As shown in Fig. \ref{fig4-2}, we also select two groups of samples from the ShanghaiTech, CUHK Avenue, and UCSD Ped2 datasets respectively for predicted frames comparison. In each group, the left column is the ground truth $I_{t+1}$. The mid column is the corresponding predicted frame $\hat{I}_{t+1}$. And the right column is the difference between predicted frame and their ground truth. 
The three groups on the right represent normal frames with almost no highlight region on their corresponding difference maps. The three groups on the left represent abnormal frames. 
As shown in Fig. \ref{fig4-2}, when there is no abnormal event, the frames can be well predicted. However, when abnormal events occur, the predictions are blurred and with color distortion. And the difference maps mainly highlight the abnormal locations in the pictures, which shows that our method can detect anomalies sensitively.

\subsection{Detailed Analysis}
In addition, we also conduct an experiment to observe the performance of our model with different memory sizes. Finally, we examine the effect of discriminative loss and compare the running time of different state-of-the-art methods.

\subsubsection{Memory Module}
 Table \ref{tab:cap3} compares AUCs of our model with different memory sizes on UCSD Ped2 and CUHK Avenue. Our method is not sensitive to the setting of memory size $N$. With a large enough memory size, our model will have a slightly better performance. Compared with MemAE \cite{gong2019memorizing}, the performance of our model with 100 memory items is still at least 3.3\% higher than that of MemAE with 2000 memory items on all datasets. Although MNAD \cite{park2020learning} uses only 10 memory items, the performance of our method is 4\% higher than it on ShanghaiTech, which is the most challenging dataset. Our model can make such great improvement because we use a prototype-guided memory module to perform discriminative latent embedding so that each individual memory item has a stronger memory ability.


\subsubsection{The Effect of Discriminative Loss}
\begin{figure}[tbp]
\centering
\includegraphics[width=0.8\linewidth]{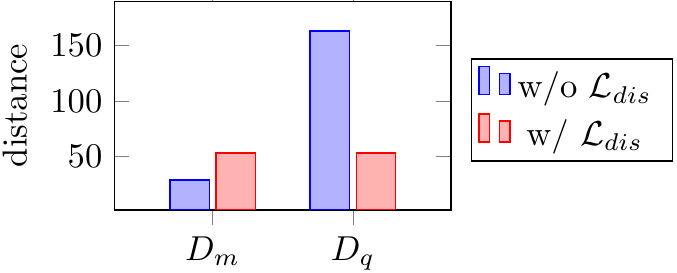}
\caption{The distance between memory items and the distance between queries and their nearest memory item.}
\label{fig7}
\vspace{-0.2cm}
\end{figure}

We calculate the distance $d(\bm{m}_j, \bm{\bar{m}})$ between each memory item $\bm{m}_j$ to the mean memory item $\bm{\bar{m}}$, and we denote the sum of these distances as:
\begin{equation}\label{eq20}
D_m = \sum_{j=1}^{N}d(\bm{m}_j, \bm{\bar{m}}),
\end{equation}
\begin{equation}
d(\bm{m}_j, \bm{\bar{m}}) = \frac{\bm{m}_j\bm{\bar{m}}^\mathsf{T}}{\lVert\bm{m}_j\lVert\lVert\bm{\bar{m}}\lVert}.
\end{equation}
At the same time, we calculate the distance between each query and its closest memory item when testing on UCSD Ped2, and we denote the average of these distances as:
\begin{equation}\label{eq21}
D_q = \frac{1}{K \times T}\sum_{t=1}^{T}\sum_{i=1}^{K}d(\bm{z}_i^t, \bm{m}_{1st}),
\end{equation}
\begin{equation}
d(\bm{z}_i^t, \bm{m}_{1st}) = \frac{\bm{z}_i^t\bm{m}_{1st}^\mathsf{T}}{\lVert\bm{z}_i^t\lVert\lVert\bm{m}_{1st}\lVert},
\end{equation}
where $t$ denote the frame index of video frames of test set with size of $T$. From Fig. \ref{fig7}, we can see that using our discriminative loss $\mathcal{L}_{dis}$ to train the model can increase the distance between memory items while making the queries closer to their nearest memory item. This shows that our loss can improve the discriminability and representativeness of memory items, so that each memory item can be regarded as a prototype.

\subsubsection{Running Time}
Our framework is implemented with NVIDIA GeForce 1080 Ti and PyTorch \cite{paszke2017automatic}. The average running time is about 42 fps on UCSD Ped2. Because we use a fast RGB difference strategy to simulate motion information, our method can achieve high detection speed and is suitable for real-time surveillance video analysis. We also report the running time of other state-of-the-art methods such as 20 fps in \cite{tudor2017unmasking}, 25 fps in \cite{liu2018future}, 32 fps in \cite{chang2020clustering}, 38 fps in \cite{gong2019memorizing}, and 67 fps in \cite{park2020learning}. The results are copied from the original corresponding papers.

\section{Conclusion}
We present a memory-augmented two-branch autoencoder for video anomaly detection. The model is designed as a combination of two branches to learn spatial and temporal information simultaneously. A novel prototype-guided memory module with discriminative loss is also introduced to perform discriminative latent embedding. Our method can learn better latent representations and thus detect anomalies sensitively. The experimental results show that our method outperforms existing state-of-the-art by a large margin.

\bibliographystyle{IEEEtran}
\bibliography{icdm21-lyd}

\end{document}